\documentclass[10pt,twocolumn,letterpaper]{article}

\usepackage{cvpr}

\usepackage{graphicx}
\usepackage{amsmath}
\usepackage{amssymb}
\usepackage{booktabs}
\usepackage{comment}
\usepackage{bm}
\usepackage{adjustbox}
\usepackage{array}
\newcolumntype{R}[2]{%
    >{\adjustbox{angle=#1,lap=1.3\width-(#2)}\bgroup}%
    l%
    <{\egroup}%
}
\usepackage{pifont}
\usepackage{stmaryrd}
\usepackage{multirow}
\usepackage{enumitem}
\usepackage{makecell}
\usepackage{setspace}
\usepackage{cuted}
\usepackage{colortbl}
\usepackage{titletoc}

\usepackage[accsupp]{axessibility}

\usepackage[pagebackref,breaklinks,colorlinks]{hyperref}

\usepackage[dvipsnames]{xcolor}

\newcommand\blfootnote[1]{%
  \begingroup
  \renewcommand\thefootnote{}\footnote{#1}%
  \addtocounter{footnote}{-1}%
  \endgroup
}

\usepackage[capitalize]{cleveref}
\crefname{section}{Sec.}{Secs.}
\Crefname{section}{Section}{Sections}
\Crefname{table}{Table}{Tables}
\crefname{table}{Tab.}{Tabs.}

\newcommand{\Rbb}{\ensuremath{\mathbb{R}}}

\newcommand{\inv}[1]{\ensuremath{\frac{1}{#1}}}

\renewcommand{\leq}{\ensuremath{\leqslant}}

\newcommand{\ma}[1]{\ensuremath{\mathsf{#1}}}
\renewcommand{\vec}[1]{\ensuremath{\bm{#1}}}

\newcommand{\abs}[1]{\ensuremath{\left| #1 \right|}}
\newcommand{\scp}[2]{\ensuremath{{\langle #1, #2 \rangle}}}
\newcommand{\set}[1]{\ensuremath{\mathcal{#1}}}

\renewcommand{\th}{\ensuremath{\text{th}}}
\def \ours{SLidR\xspace} 
\def \ppkt{PPKT$^\dagger$\xspace}
\def \depthc{DepthContrast$^\dagger$\xspace}
\def \pointc{PointContrast$^\dagger$\xspace}
\definecolor{dgreen}{rgb}{0.0, 0.5, 0.0}
\definecolor{better}{rgb}{0.19, 0.55, 0.91}
\definecolor{worse}{rgb}{0.82, 0.1, 0.26}
\definecolor{first}{rgb}{0.13, 0.67, 0.8}
\definecolor{second}{rgb}{0.74, 0.83, 0.9}
\newcommand{\cmark}{\textcolor{better}{\ding{51}}}%
\newcommand{\xmark}{\textcolor{worse}{\ding{55}}}%

\newcommand{\ra}[1]{\renewcommand{\arraystretch}{#1}}

\newcommand{\smallparagraph}[1]{\medskip\noindent\textbf{#1}~~}

\begin{document}

\def\ourtitle{Image-to-Lidar Self-Supervised Distillation for Autonomous Driving Data}
\title{\ourtitle}

\author{Corentin Sautier
\textsuperscript{1}, Gilles Puy\textsuperscript{1}, Spyros Gidaris\textsuperscript{1}, Alexandre Boulch\textsuperscript{1}, Andrei Bursuc\textsuperscript{1}, Renaud Marlet\textsuperscript{1,2}\\
\textsuperscript{1}valeo.ai, Paris, France\\
\textsuperscript{2}LIGM, Ecole des Ponts, Univ. Gustave Eiffel, CNRS, Marne-la-Vall\'ee, France\\
}

\maketitle

%
\begin{abstract}
Segmenting or detecting objects in sparse Lidar point clouds are two important tasks in autonomous driving to allow a vehicle to act safely in its 3D environment. The best performing methods in 3D semantic segmentation or object detection rely on a large amount of annotated data. Yet annotating 3D Lidar data for these tasks is tedious and costly. In this context, we propose a self-supervised pre-training method for 3D perception models that is tailored to autonomous driving data. Specifically, we leverage the availability of synchronized and calibrated image and Lidar sensors in autonomous driving setups for distilling self-supervised pre-trained image representations into 3D models. Hence, our method does not require any point cloud nor image annotations. The key ingredient of our method is the use of superpixels which are used to pool 3D point features and 2D pixel features in visually similar regions. We then train a 3D network on the self-supervised task of matching these pooled point features with the corresponding pooled image pixel features. The advantages of contrasting regions obtained by superpixels are that: (1) grouping together pixels and points of visually coherent regions leads to a more meaningful contrastive task that produces features well adapted to 3D semantic segmentation and 3D object detection; (2) all the different regions have the same weight in the contrastive loss regardless of the number of 3D points sampled in these regions; (3) it mitigates the noise produced by incorrect matching of points and pixels due to occlusions between the different sensors.
Extensive experiments on autonomous driving datasets demonstrate the ability of our image-to-Lidar distillation strategy to produce 3D representations that transfer well on semantic segmentation and object detection tasks.
\blfootnote{Code available at \href{https://github.com/valeoai/SLidR}{\color{blue}https://github.com/valeoai/SLidR}}
\end{abstract}

%
\section{Introduction}
\label{sec:intro}

\begin{figure*}[t]
	\centering
	\includegraphics[width=\linewidth]{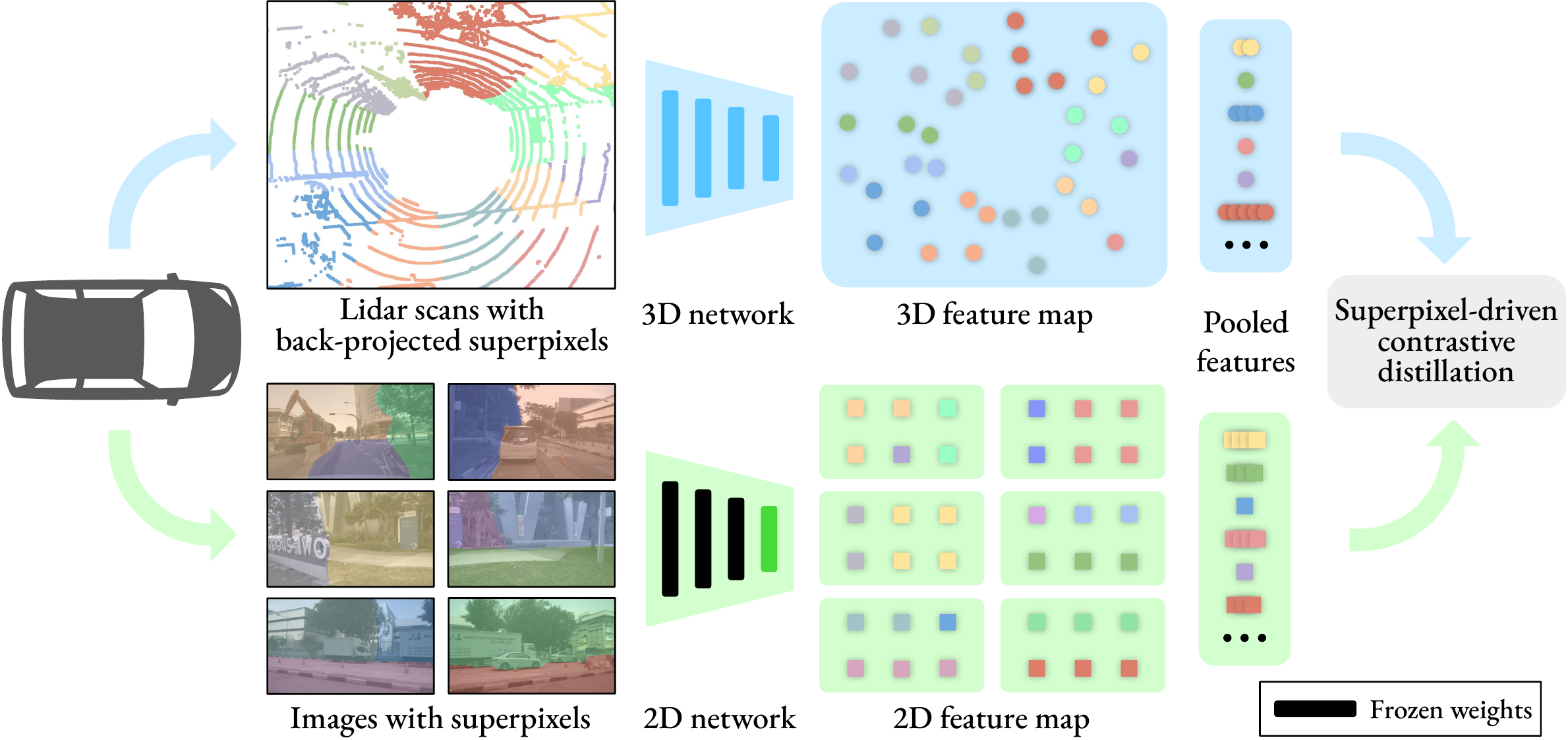}
	\caption{
	\ours distillates the knowledge of a pre-trained and fixed 2D network into a 3D network. It uses superpixels to pool features of visually similar regions together, both on the images, and on the point clouds through superpixels back-projection. The superpixel-driven contrastive loss aligns the pooled point and image features. The visualized segments proposed in this figure have been manually generated and are intentionally over-sized for illustrative purposes. Superpixels actually used can be observed on \cref{fig:slic}.} 
	\label{fig:teaser}
\end{figure*}
%

Lidar sensors deliver rich information about the 3D world, and making sense of this kind of information is crucial for an autonomous driving vehicle to properly act in its environment, across different external conditions. State-of-the-art methods for semantic segmentation or object detection in Lidar point clouds rely on deep neural networks trained on large collections of annotated point clouds. Yet, annotating 3D Lidar point clouds is a long and costly task~\cite{semantickitti,genova2021learning}. Self-supervision reduces the burden of annotating large datasets by exploiting a large amount of non-annotated data to pre-train neural networks, which are subsequently fine-tuned on a smaller set of annotated data.

The current best performing self-supervised techniques for 3D neural networks working on real point clouds are mostly adapted to indoor scenes with dense point clouds. These methods suffer from several shortcomings when dealing with sparse point clouds, such as those acquired outdoor by a moving vehicle. For example, PointContrast \cite{pointcontrast} requires pairs of registered point clouds and a list of matching points between them. While multiple reliable matching points can be found in a densely sampled static scene, the number of such pairs of points is much lower in autonomous driving datasets, in particular on objects of interest (cars, trucks, pedestrians, etc.) as they are sparsely sampled and likely to move between two acquisitions. DepthContrast \cite{depthcontrast} avoids the need of finding pairs of corresponding points as it only requires a single representation for each scene. This representation is computed by global pooling and therefore loses information on small objects. These design choices limit significantly the performance of these methods in our experiments on autonomous driving scenes.

Our goal is to design a self-supervised method for tasks such as semantic segmentation or object detection in Lidar point clouds, and tailored to autonomous driving data. 
Most autonomous driving vehicles are equipped with an array of cameras and Lidar sensors that are synchronized and calibrated, offering rich surround-view information. These data are a lot easier to acquire than to annotate, and we propose to leverage them to distill self-supervised pre-trained image representations into a 3D network.
This whole pre-training process does not require any annotation of the images nor of the point clouds. Self-supervised pre-training on images has proven very successful for learning generic representations that transfer well to various complex downstream tasks in 2D, often surpassing supervised pre-training \cite{he2019moco,swav,byol,obow}. In this work, we show that these powerful representations can also be used to pre-train 3D networks for autonomous driving. We call this setting self-supervised 2D-to-3D representation distillation.

We propose a distillation loss suited to tasks such as semantic segmentation and object detection by forcing the networks to produce object-aware representations. Inspired by \cite{detcon}, we use superpixels \cite{slic, felzenszwalb} which group visually similar regions that are likely to belong to the same object. We then use these superpixels as pooling masks for 3D point features and 2D pixel features, and enforce pairs of corresponding pooled features to match each other using a contrastive loss, as illustrated in \cref{fig:teaser}.
This pooling strategy naturally mitigates two drawbacks encountered in autonomous driving data: (1)~It reduces the noise induced by incorrect matching of points and pixels (which is performed automatically), e.g., caused by occlusions for one of the sensors; (2)~It balances asymmetries between  areas with denser coverage of points and sparser areas, that would otherwise have different weights in the contrastive loss.
The latter is particularly important for objects such as cars, pedestrians and cyclists, that are sampled more sparsely than the road near the ego-vehicle.

Finally, we examine key elements of our image-to-Lidar distillation method. This includes a careful design of the image feature projection head to avoid degenerate cases where no useful information is transferred to the 3D network.

\noindent In summary, our contributions are the following.

\begin{itemize}[itemsep=-1pt,topsep=-1pt]
\item We propose a novel self-supervised 2D-to-3D representation distillation approach based on a superpixel-to-superpoint contrastive loss and a carefully designed image feature upsampling architecture that allows high resolution image features to be distilled without suffering from degenerate solutions. 
We call this method \ours, for Superpixel-driven Lidar Representations.

\item To the best of our knowledge, this work provides the first study on the self-supervised image-to-Lidar representation distillation problem for autonomous driving data.
This includes extensively evaluating our method for the downstream tasks of semantic segmentation on nuScenes \cite{nuscenes} and SemanticKITTI \cite{semantickitti} and object detection on KITTI \cite{Geiger2012CVPR}, and comparing it against strong baselines. The latter were produced by adapting and optimizing several existing self-supervised pre-training methods for the autonomous driving setting.

\item We demonstrate that our image-to-Lidar pre-training strategy surpasses, in all evaluation settings, state-of-the-art 3D self-supervised pre-training methods and prior 2D-to-3D distillation methods, devised for dense point clouds captured in indoor scenes.

\end{itemize}

\smallskip \noindent

\section{Related Works}
\label{sec:ref}

\subsection{Self-Supervised Representation Learning}

Self-supervised methods aim to learn good representations by pre-training a neural network with an annotation-free pretext task using many unlabeled data. The goal is for these self-supervised representations to transfer well to downstream tasks of interest 
for which there are 
limited annotated data available. In the following we review recent self-supervised methods in the image and 3D domain.

\smallparagraph{2D Self-Supervision.}
Several approaches have been proposed for representation learning on the image domain~\cite{chen2020simple, doersch2015unsupervised, gidaris2018unsupervised, misra2016shuffle, noroozi2016unsupervised, pathak2016context, zhang2016colorful}.
One of the most prominent category of methods are those based on contrastive-based instance discrimination objectives~\cite{chen2020simple,he2019moco,henaff2020data,misra2020self,oord2018representation,tian2020contrastive,wu2018unsupervised,dimensionalityreduction}, which learn to match different views of the same image data (e.g., generated with random image augmentation) in the presence of distracting, negative examples.
Other prominent categories are the feature reconstruction learning methods~\cite{byol,bownet,obow,chen2021exploring,caron2021emerging} and clustering-style methods~\cite{swav,asano2019selflabelling,deepcluster,huang2019unsupervised,xie2016unsupervised,zhuang2019local}.
In this work, we exploit powerful self-supervised image representations, specifically MoCo~\cite{he2019moco, chen2020mocov2}, to pre-train 3D Lidar networks.
Apart from image-wise self-supervised objectives, as those already mentioned, there have also been proposed pixel-wise~\cite{wang2021dense, xie2021propagate, xiong2021self, van2021unsupervised} or region-wise self-supervised objectives~\cite{detcon}.
Our superpixel-driven image-to-lidar contrastive distillation loss relates to \cite{van2021unsupervised, detcon} that exploit unsupervised segmentation masks for defining their proposed contrastive objectives.

\smallparagraph{3D Self-Supervision.}
Most of the 3D self-supervised methods focus on single objects for tasks such as object recognition or part segmentation. We find techniques based on pretext tasks defined at the object level, based on point cloud reconstruction or prediction of a global transformation \cite{Sauder2019SelfSupervisedDL,Wang2020PreTrainingBC,chen2021shape,Poursaeed2020SelfSupervisedLO}.
As in 2D self-supervision, some techniques define their pretext task at the feature level using cluster prediction \cite{Hassani2019UnsupervisedMF}, contrastive-based instance discrimination methods \cite{sanghi2020info3d,wang2021unsupervised,chen2021unsupervised,du2021self}, a combination of the last two \cite{zhang_3dv_2019}, or multimodal object representations \cite{Jing_2021_CVPR}. 
Among methods working on entire scenes rather than on single objects, \cite{pointcontrast, Hou_2021_CVPR,coconets} pre-train a 3D network by learning to match points in two registered point clouds. DepthContrast \cite{depthcontrast} uses a scene-level instance discrimination pretext task both in indoor and outdoor scenes. 
Finally, \cite{Liang_2021_ICCV,huang2021spatio} both appeared publicly recently. The first presents a method extensively tested on autonomous driving scenes while the second applies on sequences of point clouds captured indoor or outdoor. Unlike us, none leverages the image modality.

\smallparagraph{Discussion.} For pre-training, self-supervised methods have relied so far on curated and balanced datasets, e.g., ImageNet~\cite{imagenet}. Self-supervised methods are highly dependent on the quality of the training data, so the dataset constitutes essentially a form of supervision in itself. In contrast, autonomous driving data, acquired from city streets, is raw and uncurated, displaying strong redundancy and imbalance. 
Here, self-supervised learning is both challenging and highly necessary to reduce the burden of continuous annotation of data, yet it hasn't been addressed much~\cite{depthcontrast,van2021revisiting}. We tackle this problem and show that we can overcome some of these challenges by leveraging multi-modality.

\subsection{Knowledge Distillation}
The purpose of knowledge distillation (KD) is to transfer useful information from a trained teacher network into a student network. To this end, the student is trained to mimic some characteristic of the teacher, 
e.g., output~\cite{hinton2015distilling} or intermediate features~\cite{romero2014fitnets, tian2020contrastive}. Initially used for distilling a large network or ensemble into a smaller network~\cite{bucilua2006model, hinton2015distilling, korattikara2015bayesian}, KD has been recently revisited as teacher-student architectures for semi-supervised~\cite{laine2017temporal, tarvainen2017mean} and unsupervised representation learning~\cite{bownet,obow,caron2021emerging,byol}. Here, after training the student typically outperforms the teacher.

\smallparagraph{2D-to-3D knowledge distillation.}
Our work relates to the setting of KD from a 2D teacher pre-trained on images into a 3D student network~\cite{jaritz2020xmuda, gupta2016cross, ppkt}. For instance, in \cite{gupta2016cross} indoor RGB-D data is used for distilling an RGB teacher into a 2D student network for depth-maps.
The most related to ours is the unpublished concurrent work~\cite{ppkt} where 2D representations are distilled in a 3D network. 
Besides being designed for dense indoor RGB-D data, a major difference between the two methods is that \cite{ppkt} contrasts pixels with points while our method contrasts 2D image regions with 3D point cloud regions, defined using superpixels. 
We explain the advantages of this superpixel-based distillation formulation in \cref{sec:superpixel_contrastive_loss}.
Moreover, in absence of public code, we developed and optimized our best adaptation of this method for autonomous driving data and our empirical comparison with it in \cref{sec:exp} demonstrates the superiority of our method.
Finally, the idea of contrasting point-pixel pairs is also exploited in \cite{p4contrast} and \cite{pri3d} for 2D-3D modality fusion and building geometry-aware 2D networks, respectively, rather than for KD from a 2D network to a 3D network as done in our method.

\section{Our approach}
\label{sec:method}

\begin{figure}[t]
	\centering
	\includegraphics[width=\linewidth]{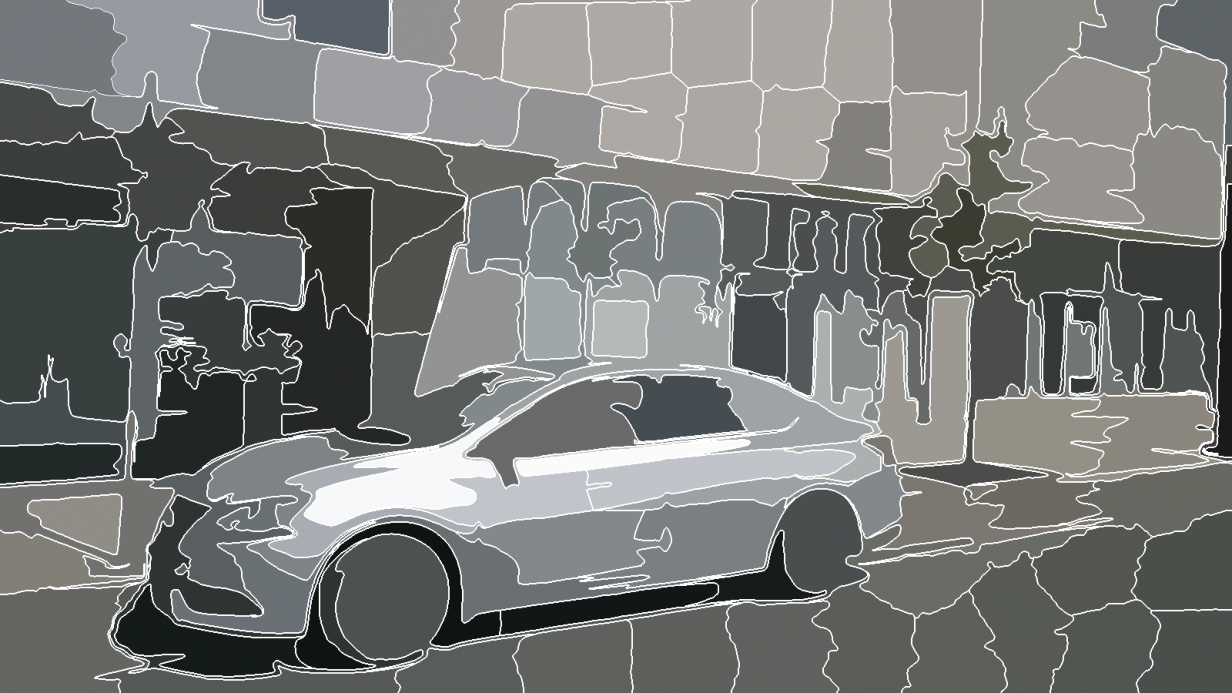}
	\vspace{-20pt}
	\caption{SLIC superpixels computed on an image from nuScenes}
	\label{fig:slic}
\end{figure}

\subsection{Image-to-Lidar Self-supervised Distillation} \label{sec:overview}

Our goal is to learn, by self-supervised distillation, 3D Lidar representations by leveraging the availability of aligned Lidar and image data in autonomous driving setups.

\smallparagraph{Synchronized Lidar and image data.} 
Let $\ma{P} = (\vec{p}_i)_{i=1, \ldots, n} \in \Rbb^{N \times 3}$ denote a point cloud captured in a scene by a Lidar at time $t_0$. 
We assume that $C$ color images $\ma{I}_1, \ldots, \ma{I}_C \in \Rbb^{M \times 3}$, where $M$ denotes the number of pixels, are captured by $C$ cameras in the same scene at $t_0$ and that the relative poses between the Lidar and cameras sensors are known. 
The pose information permits us to project each point $\vec{p}_i$ in the $C$ camera frames.
Specifically, we can build a mapping $\rho_c \colon \Rbb^3 \rightarrow \{1, \ldots, M\} \cup \{ 0 \}$, for each camera $c$, that takes as input a 3D point $\vec{p}_i$ and outputs the index of the corresponding 2D pixel in the frame $\ma{I}_c$, or $0$ if the input point is not viewed in camera $c$. 

\smallparagraph{Distilling network representations.}
Let $f_{\theta_{\scriptscriptstyle \text{bck}}}\colon \Rbb^{N \times 3} \rightarrow \Rbb^{N \times D}$ be a 3D deep neural network, with trainable parameters $\theta_{\scriptscriptstyle \text{bck}}$, that takes as input a point cloud and outputs one $D$-dimensional feature per point. 
Our goal is to pre-train this 3D network without using any human annotations.
To this end, we exploit the aligned and synchronized Lidar and image data, described in the previous paragraph. We also leverage the availability of a self-supervised pre-trained image network $g_{\bar{\omega}_{\scriptscriptstyle \text{bck}}}\colon \Rbb^{M \times 3} \rightarrow \Rbb^{M' \times E}$ with trained and fixed parameters $\bar{\omega}_{\scriptscriptstyle \text{bck}}$, that takes as input an image and outputs an $E$-dimensional feature map at a possibly lower resolution $M' \leq M$.
In this context, we propose to train $f_{\theta_{\scriptscriptstyle \text{bck}}}(\cdot)$ by aligning the point features $f_{\theta_{\scriptscriptstyle \text{bck}}}(\ma{P})$ with the pre-trained image representations $g_{\bar{\omega}_{\scriptscriptstyle \text{bck}}}(\ma{I}_1), \ldots, g_{\bar{\omega}_{\scriptscriptstyle \text{bck}}}(\ma{I}_C)$. We achieve this goal with a superpixel-driven contrastive loss, which we describe in the following section.

\subsection{Superpixel-driven Contrastive Distillation Loss} \label{sec:superpixel_contrastive_loss}

To distill the knowledge of the pre-trained image network $g_{\bar{\omega}_{\scriptscriptstyle \text{bck}}}(\cdot)$ into the 3D Lidar network $f_{\theta_{\scriptscriptstyle \text{bck}}}(\cdot)$ by self-supervision, 
we use a contrastive loss~\cite{oord2018representation} between the image features obtained from $g_{\bar{\omega}_{\scriptscriptstyle \text{bck}}}(\cdot)$ and the 3D point features extracted from $f_{\theta_{\scriptscriptstyle \text{bck}}}(\cdot)$. 
As we are interested in downstream tasks such as semantic segmentation and object detection, the learned 3D representations should ``reason'' in terms of objects or object parts. 
We want to contrast features at the object level rather than at the over-detailed pixel level or the overly-coarse scene level.

To that end, we use superpixels for grouping pixels which are locally visually similar, hence likely to belong to one object, and we define our contrastive loss with them. 
We segment the image $\ma{I}_c$ into, at most, $Q$ superpixels with SLIC \cite{slic}, as illustrated in \cref{fig:slic}. We denote the superpixels by $\mathcal{S}_1^c, \ldots, \mathcal{S}_Q^c$, where $\mathcal{S}_s^c$ is the set of pixel indices belonging to the $s^\th$ superpixel.
We have $\mathcal{S}_1^c \cup \ldots \cup \mathcal{S}_Q^c = \{1, \ldots, M\}$ and $\forall s \neq s',\ \mathcal{S}_s^c \cap \mathcal{S}_{s'}^c = \varnothing$. 
We also use the mapping function $\rho_c$ to group the points viewed in the $c^\th$ camera into $Q$ distinct superpoints: $\mathcal{G}_1^c, \ldots, \mathcal{G}_Q^c$, where $\mathcal{G}_s^c = \{ i : \rho_c(\vec{p}_i) \in \mathcal{S}_s^c\}$. In the ideal case of one object per superpixel, we want the point feature describing $\mathcal{G}_s^c$ to be similar to the image feature describing the corresponding superpixel $\mathcal{S}_s^c$, 
but unlike the image feature describing a different superpixel $\mathcal{S}_{s'}^{c'}$ from the same scene, for $(s, c) \neq (s', c')$, or the image feature describing a superpixel from a different scene.
This distillation principle is formalized as follows.

\smallparagraph{Contrastive loss.} 
For each camera $c$, we compute superpoint and superpixel features by average pooling:
\begin{align}
\label{eq:superpoint_feature}
\vec{f}^c_s &= \inv{\abs{\mathcal{G}_s^c}} \sum_{i \in \mathcal{G}_s^c} (h_{\theta_{\scriptscriptstyle \text{head}}} \circ f_{\theta_{\scriptscriptstyle \text{bck}}})(\ma{P})_i
\\
\label{eq:superpixel_feature}
\vec{g}^c_s &= \inv{\abs{\mathcal{S}_s^c}} \sum_{j \in \mathcal{S}_s^c} (h_{\omega_{\scriptscriptstyle \text{head}}} \circ g_{\bar{\omega}_{\scriptscriptstyle \text{bck}}})(\ma{I}_{c})_j,
\end{align}
for all $s$ such that $\abs{\mathcal{G}_s^c} > 0$ (a superpixel feature is computed only if the corresponding superpoint is non-empty). The heads $h_{\theta_{\scriptscriptstyle \text{head}}}, h_{\omega_{\scriptscriptstyle \text{head}}}$, with trainable parameters $\theta_{\scriptscriptstyle \text{head}}$ and $\omega_{\scriptscriptstyle \text{head}}$, project 
respectively the 3D-based and 2D-based features into the same
$F$-dimensional space. The point projection head $h_{\theta_{\scriptscriptstyle \text{head}}} \colon \Rbb^{N \times D} \rightarrow \Rbb^{N \times F}$ is a simple pointwise linear layer followed by $\ell_2$-normalization. The pixel projection head $h_{\omega_{\scriptscriptstyle \text{head}}} \colon \Rbb^{M' \times E} \rightarrow \Rbb^{M \times F}$ is described in \cref{sec:technical_details}.

We transfer the knowledge from the 2D network to the 3D network by using a contrastive loss which favors a solution where a superpoint feature $\vec{f}^c_s$ is more correlated to its corresponding superpixel features $\vec{g}^c_s$ than any other feature $\vec{g}^{c'}_{s'}$, with $(s, c) \neq (s', c')$. Concretely, the network $f_{\theta_{\scriptscriptstyle \text{bck}}}$ and the projection heads $h_{\theta_{\scriptscriptstyle \text{head}}}, h_{\omega_{\scriptscriptstyle \text{head}}}$ are jointly trained using the following superpixel-driven contrastive loss:
\begin{align}
\label{eq:distillation_loss}
\set{L}(\theta_{\scriptscriptstyle \text{bck}}, \theta_{\scriptscriptstyle \text{head}}, \omega_{\scriptscriptstyle \text{head}}) 
=
-\sum_{c, s}\log\left[ 
\frac
{\exp(\scp{\vec{f}^c_s}{\vec{g}^c_s} / \tau)}
{\sum\limits_{c',s'} \exp(\scp{\vec{f}^c_s}{\vec{g}^{c'}_{s'}} / \tau)}
\right],
\end{align}
where $\scp{\cdot}{\cdot}$ denotes the scalar product in $\Rbb^F$, and $\tau > 0$ a temperature. For simplicity, we have only defined here the contrastive loss for a single scene. In practice, multiple scenes are available in a batch at training time and the denominator in Eq.\,\eqref{eq:distillation_loss} actually contains the superpixel features of \emph{all} scenes in the batch.

\smallparagraph{Discussion.} We argue that using a contrastive loss at the superpoint-superpixel level is better adapted for pre-training $f_{\theta_{\scriptscriptstyle \text{bck}}}$ for semantic segmentation and object detection than staying at the point-pixel level. 

First, as already mentioned, superpixels permit us to group points and pixels in visually similar regions, thus likely to belong to one object. 
Hence, the loss \eqref{eq:distillation_loss} will favor point features with the desirable property of being locally coherent when they belong to the same object (assuming the superpixel does not cover several objects). 
Furthermore, compared to contrasting pixels to points, we lower the proportion of ``false negatives" in the contrastive loss as we do not contrast between almost identical points inside a superpixel. Yet, a small number of those false negatives can remain as the superpixels tend to oversegment the objects, but this is common in unsupervised contrastive learning where similar or same class images can be contrasted due to the instance discrimination setting. In addition, our strategy of averaging features within superpixels limits the impact of pixels belonging to different semantic regions.
On the other end of the spectrum, scene-level contrastive learning, i.e., contrasting the global representation of an entire point cloud to the global representation(s) of the corresponding camera frames, is not meaningful for autonomous driving data: (a) autonomous driving scenes consist of multiple different objects; (b) there is relatively limited diversity at the scene-level since all scenes consist of almost the same type of objects, e.g., roads, cars, and pedestrians.

Second, superpixels permit us to naturally give the same weights to all regions in the contrastive loss irrespective of the point sampling density in those regions. 
In typical Lidar scans from AD scenes, the density of points varies greatly with a majority of points sampled on the road and near the ego-vehicle.
In the distillation loss of Eq.\,\eqref{eq:distillation_loss} 
if we consider points and pixels instead of superpoints and superpixels, we cannot in practice exhaustively consider all possible pairs of matching point and pixel because of computational tractability, and we must subsample them, as done in~\cite{pointcontrast,ppkt}.
However, if this subsampling is performed randomly, without any proper selection, the loss in Eq.\,\eqref{eq:distillation_loss} is dominated by points in high density regions. 
With superpixel-based pooling of point features, we reduce the number of pairs, thus removing the need for subsampling, and the loss now treats different objects equally, whether they are seen in a region sampled with high or low density.

Third, the Lidar and image sensors have different viewpoints and the respective acquisitions are never perfectly synchronized. Hence, the point-pixel matching is only approximate, with incorrect matches
due to sensor occlusions and motion.
Averaging the features as in Eq.~\eqref{eq:superpoint_feature} and Eq.~\eqref{eq:superpixel_feature} allows us to reduce the impact of spurious matches. 

\subsection{The Devil is in the Image Projection Head}
\label{sec:technical_details}

In our method, the image network $g_{\bar{\omega}_{\scriptscriptstyle \text{bck}}}(\cdot)$ is a ResNet-50 pre-trained under self-supervision on ImageNet~\cite{imagenet} using MoCov2~\cite{he2019moco,chen2020mocov2}. This network outputs features at a much lower resolution than the resolution of input images: $M' = M / 32^2$. To recover features at the pixel level and be able to compute Eq.\,\eqref{eq:superpixel_feature}, we tested several architectures for the pixel projection head $h_{\omega_{\scriptscriptstyle \text{head}}}$ and were able to reach good performance only when using a $1 \times 1$ convolution layer followed by a fixed upsampling method, such as bilinear or nearest neighbor upsampling.
Indeed, we noticed that having a $h_{\omega_{\scriptscriptstyle \text{head}}}$ architecture that captures wide spatial context for an output pixel feature (e.g., using several convolutional layers with kernel size greater than 1) allows $h_{\omega_{\scriptscriptstyle \text{head}}}$ to ``cheat'' on the contrastive task with the ``leaked'' context information, by matching a 2D image feature with its paired 3D point(s) only based on the 2D spatial position of the 2D feature inside the image.
This would allow the network to solve the contrastive task, however it is of no use for learning high-level semantic features.
We expect similar degenerate solutions if $g_{\bar{\omega}_{\scriptscriptstyle \text{bck}}}(\cdot)$ is not kept frozen.

Empirically, we noticed that performance improves as we preserve the input resolution in the ResNet-50 encoder.
We propose a few simple adjustments to the ResNet-50 encoder to limit downsampling without increasing significantly the computational complexity or adding new parameters. We keep the first strided convolution and max pooling layer, but then maintain a fixed resolution across all residual blocks by replacing strided convolutions with dilated convolutions.
This leads to the lesser discrepancy $M' = M / 4^2$. 

To conclude, to bridge the gap between the two resolutions, the pixel projection head $h_{\omega_{\scriptscriptstyle \text{head}}}$ consists of a pixel-wise convolution, followed by a bilinear upsampling layer by $4$ in each spatial direction, and an $\ell_2$-normalization layer.

\section{Experiments} \label{sec:exp}

\subsection{3D Network Pre-training}

\smallparagraph{Network backbones.} 
The 3D backbone $f_{\theta_{\scriptscriptstyle \text{bck}}}$ is the sparse residual U-Net architecture used in \cite{pointcontrast}, which was originally designed in \cite{4dspatio}, where we use $3\times3\times3$ kernels for all sparse convolutions. 
As input representation it takes a sparse occupancy grid of the 3D data obtained by quantizing the 3D points, i.e., voxels.
Instead of using voxel in Cartesian coordinates as in \cite{pointcontrast,depthcontrast,ppkt}, we use voxels in cylindrical coordinates which are better suited for Lidar point clouds \cite{cylinder3D}. 
The voxel size that we use is $10$ cm on the $z$-axis and radius (distance to the origin in the $xy$-plane) component, and $1^\circ$ on the azimuth angle.
The 2D backbone is a ResNet-50~\cite{resnet} pre-trained with MoCov2~\cite{he2019moco,chen2020mocov2}.

\smallparagraph{Pre-training dataset.} We pre-train all models on nuScenes \cite{nuscenes}. 
This dataset contains $700$ training scenes from which we keep aside $100$ scenes that constitute our mini-val split, used to choose our training hyperparameters and do our ablation study. 
The models are pre-trained using all the keyframes from the $600$ remaining training scenes.

\smallparagraph{Training parameters.} Unless mentioned otherwise, $f_{\theta_{\scriptscriptstyle \text{bck}}}$, $h_{\theta_{\scriptscriptstyle \text{head}}}$ and $h_{\omega_{\scriptscriptstyle \text{head}}}$ are trained on $1$ GPU with \ours for $50$ epochs using SGD with a initial learning rate of $0.5$, a momentum of $0.9$, weight decay of $0.0001$, dampening of $0.1$, and a cosine annealing scheduler that decreases the learning rate from its initial value to $0$ at the end of the $50^\th$ epoch.

\smallparagraph{Data augmentation.} A key factor in the success of self-supervision is the use of strong data augmentations~\cite{depthcontrast,chen2020mocov2,obow}. We apply several augmentations detailed in the supplementary material. In summary, on the point cloud side, we apply a random rotation around the $z$-axis, flip randomly the direction of the $x$ and $y$-axis, and drop points that lie in a random cuboid as in \cite{depthcontrast}. On the image side, we use a random horizontal flip and a random crop-resize.

\subsection{Baselines}

\begin{table}[t]
\centering
\ra{1.1}
\setlength{\tabcolsep}{4pt}
\begin{tabular}{@{}l c c l l l@{}}
\toprule
    Method
        & \multicolumn{1}{l}{Dil. Conv.}
        & \multicolumn{1}{l}{Superpix.}
        & mIoU \\
\midrule
    \ppkt \cite{ppkt}
        & \xmark    
        & \xmark 
        & 34.7
        \\
    \ours w/o superpix.
        & \cmark    
        & \xmark 
        & 36.6 \textcolor{better}{(+1.9)}
        \\
    \ours
        & \cmark 
        & \cmark 
        & \bf 39.2 \textcolor{better}{(+4.5)}
        \\
\bottomrule
\end{tabular}
\vspace{-6pt}
\caption{Ablation study on nuScenes semantic segmentation by replacing \eqref{eq:distillation_loss} with a pixel-level contrastive loss (\ours w/o superpix.) and then using strided convolution instead of dilated convolution in ResNet-50 (\ppkt). The scores are obtained by linear probing of the pre-trained backbone. We report the mIoU on our mini-val split.}
\label{tab:abalation}
\end{table}

To the best of our knowledge, this is the first work to study image-to-Lidar self-supervised representation distillation on autonomous driving data. 
Hence, we cannot rely on existing baselines trained in this setup. 
In order to fairly compare against strong baselines, a significant amount of work was done to adapt and optimize existing pre-training methods to our setup.
In particular, we use 3 representative methods for 3D network pre-training as baselines: PPKT~\cite{ppkt}, PointContrast~\cite{pointcontrast}, and DepthContrast~\cite{depthcontrast}.

\smallparagraph{PPKT for autonomous driving.} 
The first baseline is PPKT, which is also a 2D-to-3D representation distillation method, but based on a pixel-to-point contrastive loss. Since PPKT was originally proposed for RGB-D data captured indoor and, up to now, there is no publicly released code,  we propose our best adaption of this method in an autonomous driving setup, referred as \ppkt.

\begin{table}[t]
\centering
\ra{1.1}
\setlength{\tabcolsep}{3pt}
\begin{tabular}{@{}l l l l@{}}
\toprule
    \multirow{3}{*}{Initialization of $f_{\theta_{\scriptscriptstyle \text{bck}}}$}
        & \multicolumn{2}{c}{nuScenes \cite{nuscenes}} 
        & KITTI \cite{semantickitti}
        \\
\cmidrule(lr){2-3} \cmidrule(lr){4-4}
        & \multicolumn{1}{l}{Lin. prob.} 
        & \multicolumn{1}{l}{Finetuning}  
        & \multicolumn{1}{l}{Finetuning}
        \\
        & \multicolumn{1}{l}{100$\%$} 
        & \multicolumn{1}{l}{1$\%$}  
        & \multicolumn{1}{l}{1$\%$}  
        \\
\midrule
    Random 
        & \textcolor{white}{0}8.1
        & 30.3    
        & 39.5
        \\
    \pointc \cite{pointcontrast}
        & 21.9
        & 32.5 \textcolor{better}{(+2.2)} 
        & 41.1 \textcolor{better}{(+1.6)}     
        \\
    \depthc \cite{depthcontrast}
        & 22.1
        & 31.7 \textcolor{better}{(+1.9)}    
        & 41.5 \textcolor{better}{(+1.2)}     
        \\
    \ppkt \cite{ppkt}
        & 36.4
        & 37.8 \textcolor{better}{(+7.5)}
        & 43.9 \textcolor{better}{(+4.4)}      
        \\
    \ours
        & \bf 38.8
        & \bf 38.3 \textcolor{better}{(+8.0)}
        & \bf 44.6 \textcolor{better}{(+5.1)}
        \\
\bottomrule
\end{tabular}
\vspace{-6pt}
\caption{Comparison of different pre-training methods for semantic segmentation by linear probing or finetuning. On nuScenes \cite{nuscenes}, we use either $1\%$ or $100\%$ of the annotated training scans. On SemanticKITTI \cite{semantickitti}, we use $1\%$ of the annotated training scans. We report the mIoU on the validation set of nuScenes and on the sequence $8$ of SemanticKITTI.}
\label{tab:semseg}
\end{table}

\smallparagraph{PointContrast.} 
The second baseline is PointContrast \cite{pointcontrast}, which, rather than image-to-lidar distillation, learns 3D representations with a contrastive task defined only at the level of 3D points. We retrained PointContrast on nuScenes after studying several setups to optimize its performance. This method requires pairs of point clouds acquired from different viewpoints in the same scene with a list of matching points in these two views. We use pairs of nuScenes' keyframes in each scene and provide the details of the construction of this dataset in the supplementary material. We denote this adapted version \pointc.

\smallparagraph{DepthContrast.} 
The last baseline is DepthContrast \cite{depthcontrast} which pre-trains simultaneously two 3D network backbones, 
a point-based network, e.g., \cite{pointnet2} and a voxel-based network, e.g., \cite{voxelnet}, using a contrastive task between the global point-cloud representations of the two networks.
To make it comparable with the rest of the methods, we used the point-based network of \cite{depthcontrast} and used our sparse residual U-Net that processes occupancy map in cylindrical coordinate as the voxel-based network. We only evaluate the performance of this voxel-based network after pre-training.

\smallparagraph{Training setup.} All the baselines are pre-trained on the same nuScenes split as for \ours, using $1$ GPU, and after tuning the hyperparameters for our min-val split, 
except for DepthContrast for which the performance are significantly improved when using 4 GPUs.
We use the same data augmentations, voxel-based 3D network backbone, and cylindrical coordinate voxels as in our method. More implementation details are provided in the supplementary material.

\subsection{Transfer on Semantic Segmentation}

We study the quality of the learned representations obtained with \ours and the baselines for semantic segmentation.
We test the performance of these methods on nuScenes \cite{nuscenes}, the dataset viewed during pre-training, but also test the robustness to a domain change by using SemanticKITTI \cite{semantickitti}.
There are $16$ semantic classes in nuScenes and $19$ in SemanticKITTI. Except otherwise mentioned, the quality of the models is evaluated on the original validation split of nuScenes and on the sequence $8$ of SemanticKITTI. 

\begin{table}[t]
\centering
\ra{1.1}
\setlength{\tabcolsep}{9pt}
\begin{tabular}{@{}l l l l l l@{}}
\toprule
    Init. of $f_{\theta_{\scriptscriptstyle \text{bck}}}$
        & 1$\%$ 
        & 5$\%$ 
        & 10$\%$ 
        & 25$\%$ 
        & 100$\%$
        \\
\midrule
    Random      
        & 30.3 \vspace{0.3em}
        & 47.7  
        & 56.6   
        & 64.8 
        & \makecell{74.2}
        \\
    \ours       
        & \setstretch{0.6}\makecell{\bf 39.0 \\ \scriptsize \textcolor{better}{(+8.7)}}
        & \setstretch{0.6}\makecell{\bf 52.2 \\ \scriptsize \textcolor{better}{(+4.5)}}
        & \setstretch{0.6}\makecell{\bf 58.8 \\ \scriptsize \textcolor{better}{(+2.2)}}
        & \setstretch{0.6}\makecell{\bf 66.2 \\ \scriptsize \textcolor{better}{(+1.4)}}
        & \setstretch{0.6}\makecell{\bf 74.6 \\ \scriptsize \textcolor{better}{(+0.4)}}
        \\
\bottomrule
\end{tabular}
\vspace{-6pt}
\caption{Improvement of the performance thanks to \ours for semantic segmentation over a randomly initialized network as a function of the percentage of available annotations on nuScenes. We report the mIoU on the validation set of nuScenes.}
\label{tab:semseg_perf_vs_percent}
\end{table}

\smallparagraph{Evaluation settings.}
We use two evaluation protocols. 
In both, we adapt the pre-trained 3D backbones $f_{\theta_{\scriptscriptstyle \text{bck}}}$ for the semantic segmentation task by adding a point-wise linear classification head on their output.
The first protocol evaluates the quality of the pre-trained features as they are, by linear probing them.
To that end, we only train the added classification head on the nuScenes \cite{nuscenes} dataset, while keeping the pre-trained parameters of $f_{\theta_{\scriptscriptstyle \text{bck}}}$ fixed.
The second protocol evaluates the ability of the pre-trained 3D representations to learn to perform semantic segmentation in a regime where only a small number of annotations is available. 
Therefore, in this protocol, we fine-tune the entire network on the semantic segmentation task on nuScenes or semanticKITTI \cite{semantickitti} using only a portion of the available annotations.
In both protocols, we use a linear combination of the cross-entropy and the Lov\'asz loss \cite{lovasz} as training objective. For few-shot semantic segmentation, we optimized the fine-tuning learning rate for each method using our mini-val split for nuScenes and 10 percent of the training set of semanticKITTI, which remained unused during few-shot fine-tuning.
Training details are provided in the supplementary material.

\subsubsection{Ablation Study}

\begin{figure*}[t]
	\centering
	\begin{minipage}{2mm}
	    \rotatebox{90}{\scriptsize Im. feat. sim. map}
	\end{minipage}
	\begin{minipage}{0.24\linewidth}
	    \centering 
	    \includegraphics[width=\linewidth]{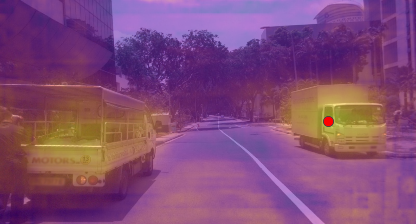}
	\end{minipage}
	\begin{minipage}{0.24\linewidth}
	    \centering 
	    \includegraphics[width=\linewidth]{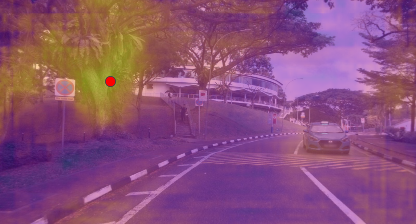}
	\end{minipage}
	\begin{minipage}{0.24\linewidth}
	    \centering 
	    \includegraphics[width=\linewidth]{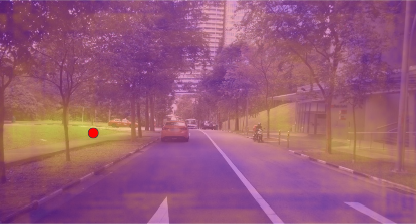}
	\end{minipage}
	\begin{minipage}{0.24\linewidth}
	    \centering 
	    \includegraphics[width=\linewidth]{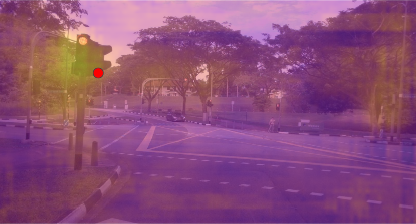}
	\end{minipage}
	\\ \vspace{0.05cm}
	\begin{minipage}{2mm}
	    \rotatebox{90}{\scriptsize Point feat. sim. map}
	\end{minipage}
	\begin{minipage}{0.24\linewidth}
	    \centering
	    \includegraphics[width=\linewidth]{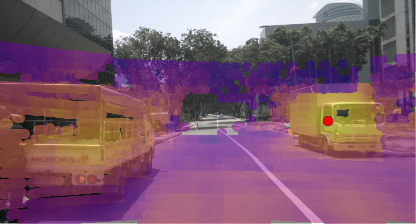}
	\end{minipage}
	\begin{minipage}{0.24\linewidth}
	    \centering
	    \includegraphics[width=\linewidth]{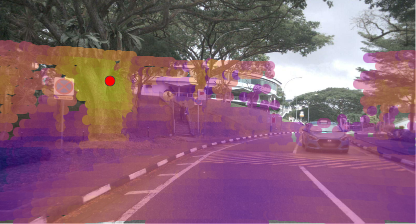}
	\end{minipage}
	\begin{minipage}{0.24\linewidth}
	    \centering
	    \includegraphics[width=\linewidth]{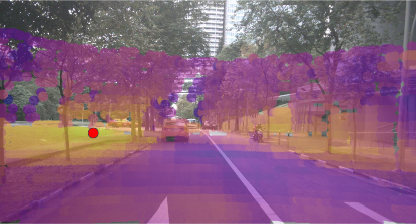}
	\end{minipage}
	\begin{minipage}{0.24\linewidth}
	    \centering
	    \includegraphics[width=\linewidth]{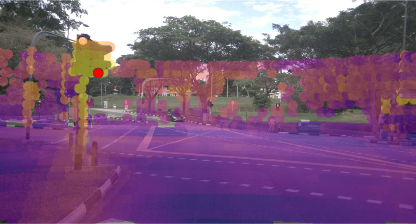}
	\end{minipage}
	\caption{
	We present the cosine similarity between the \ours's feature of a query point (displayed as a red dot) and: (a) the pixel features of an image in the same scene (top row - Image feature similarity map); (b) the features of the other points projected in the same image (bottom row - Point feature similarity map). The colormap goes from violet to yellow for respectively low and high similarity scores. We show these maps for two scenes in the validation set of \cite{nuscenes}.
	}
	\label{fig:feature_maps}
\end{figure*}

We first justify the use of dilated convolutions in the ResNet-50 and of our superpixel-driven contrastive loss \eqref{eq:distillation_loss}.
To that end, we train $3$ networks with: 
(a) \ours, 
(b) \ours without superpixels (i.e., using a point-to-pixel version of \eqref{eq:distillation_loss}), and finally
(c) \ours without superpixels and without the dilated convolution, which is our implementation of \ppkt. 
We then test the quality of the trained backbone by linear probing for semantic segmentation on nuScenes.

The results computed on our mini-val split are reported in \cref{tab:abalation} and show that each of our technical contributions improve the performance. We notice a gain of $1.9$ point thanks to the dilated convolutions and, most importantly, a huge improvement of $4.5$ point when combined with our key ingredient: the superpixel-driven contrastive loss \eqref{eq:distillation_loss}.

A complementary analysis of the sensitivity of SLidR to the choice of the underlying superpixel algorithm is available in the supplementary material.

\subsubsection{Comparisons with Baselines}

We compare \ours with the baselines on the linear probing setup on nuScenes and the few-shot end-to-end fine-tuning setup using $1\%$ of the available annotations on nuScenes and SemanticKITTI. The results are reported in \cref{tab:semseg}.

We observe that: 
\textbf{(1)} All pre-training methods are better than random initialization.
\textbf{(2)} \ppkt and our \ours, the methods based on image-to-Lidar distillation for pre-training, perform significantly better than the \depthc and \pointc approaches.
This highlights the advantage of exploiting self-supervised image pre-trained networks for learning 3D Lidar representations, as we advocate in our work.
\textbf{(3)} \ours achieves better semantic segmentation performance, especially on the linear probing setup, which demonstrates the superiority of our superpixel-driven method.

\subsubsection{Annotation Efficiency on nuScenes}

\begin{table}[t]
\centering
\ra{1.1}
\setlength{\tabcolsep}{9pt}
\begin{tabular}{@{}l l l l@{}}
\toprule
    Init. of $f_{\theta_{\scriptscriptstyle \text{bck}}}$
        & 5$\%$ 
        & 10$\%$ 
        & 20$\%$
        \\
\midrule
    Random      
        & 56.1 
        & 59.1 
        & 61.6 
        \\
    \ppkt     
        & \bf 57.8 \textcolor{better}{(\bf+1.7)} 
        & 60.1 \textcolor{better}{(+1.0)} 
        & 61.2 \textcolor{worse}{(-0.4)}
        \\
    \ours  
        & \bf 57.8 \textcolor{better}{(+1.7)} 
        & \bf 61.4 \textcolor{better}{(+2.3)} 
        & \bf 62.4 \textcolor{better}{(+0.8)} 
        \\
\bottomrule
\end{tabular}
\vspace{-6pt}
\caption{Performance of pre-training methods for object detection by fine tuning pre-trained networks using different percentages of the annotated scans in the KITTI 3D object detection dataset \cite{Geiger2012CVPR}. Scores are average mAP across cars, pedestrians and cyclists.}
\label{tab:obect_detection_perf_vs_percent}
\end{table}

We continue by studying the performance of \ours compared to a training from a random initialization as function of the percentage of annotated data for semantic segmentation on nuScenes. 
We fine-tune end-to-end the two examined networks using $1\%$, $5\%$, $10\%$, $25\%$ or $100\%$ of the annotations on nuScenes, with learning rates optimized on our mini-val split for each method and subset size.

The results are presented in \cref{tab:semseg_perf_vs_percent} and show a constant improvement over a training from a randomly initialized backbone. 
The improvement is 0.4 point on the mIoU for $100\%$ annotations and increases to up to 8.7 point as the percentage of available annotations decreases.

\subsection{Transfer for Few-Shot Object Detection}
\label{sec:few_shot_object_detect}

Here we evaluate the quality of our pre-trained Lidar representations on the challenging downstream tasks of 3D object detection on the KITTI dataset \cite{Geiger2012CVPR}.

\smallparagraph{Experimental setup.} We use OpenPCDet \cite{openpcdet} in which we modify the PointRCNN model by replacing the PointNet++ \cite{pointnet2} backbone with our pre-trained backbone. This model is fine-tuned on subsets of different sizes of the standard KITTI object detection dataset \cite{Geiger2012CVPR}, which contains bounding boxes for cars, cyclists and pedestrians. The performance are compared by computing the average mAP over these three classes in the moderately difficult cases, which are used to rank all methods on this dataset \cite{Geiger2012CVPR}.

\smallparagraph{Fine-tuning protocol.} We initialize the pre-trained backbone $f_{\theta_{\scriptscriptstyle \text{bck}}}$ with \ours, \ppkt, or using random weights. For this experiments, $f_{\theta_{\scriptscriptstyle \text{bck}}}$ have been pre-trained using 4 GPUs, a batch size of $16$, a initial learning rate of $2$, and synchronized batch-norm layers for \ppkt and \ours. 
The networks are then fine-tuned on 4 GPUs with a batch size of $12$ and the default settings of OpenPCDet. The learning rate for fine-tuning is optimized for each subset of the KITTI object detection dataset so as to maximize the performance of the backbone initialized with random weights. We then use the same learning rate for the other methods.

\smallparagraph{Results.} The results are reported in \cref{tab:obect_detection_perf_vs_percent}. We remark that \ours gives a significant improvement of up to $2.3$ points in mAP over a situation where no pre-training is done. \ours also outperforms \ppkt with a gain of at least $1.2$ point in mAP at $10\%$ and $20\%$ of annotated data.

\subsection{Visual Inspection}

The feature similarity maps presented in \cref{fig:feature_maps} highlights our pre-trained model's object recognition and segmentation abilities, by showing how features are locally coherent when they belong to the same object.
In the leftmost scene, the query point on the truck on the right side is highly correlated with points on the same truck as well as with points on the truck on the left side. 
This indicates that our self-supervised 3D features already allows distinction of objects without fine-tuning.
The same phenomenon is observed with points on the traffic lights in the rightmost scene. 
Furthermore, the nearly identical image and point feature similarity maps illustrate the quality of the knowledge transfer. Finally, we remark some spurious correlations on the road, which indicate that \ours might still be improved.

\subsection{Technical Limitations}

A first limitation might occur in low-light conditions as the computed superpixels might provide irrelevant object segments, impairing the performance of our method.

Another limitation occurs when the output image features are similar between two superpixels, e.g, $\vec{g}_{1}^c \approx \vec{g}_{2}^c$, as then, the contrastive loss will try to enforce a solution where the superpoint feature $\vec{f}_{1}^c$ is correlated to $\vec{g}_{1}^c$ but uncorrelated to $\vec{g}_{2}^c$, which is impossible.
While this issue is common in contrastive self-supervised methods, the impact is possibly a bit stronger here as the whole image backbone $g(\cdot)$ is frozen, leaving less room for adjustments in these situations.
Addressing this limitation is left for future work.

%
\section{Conclusion}
\label{sec:conclusion}

We proposed \ours, a self-supervised image-to-Lidar distillation method working on synchronized Lidar and camera data, as typically found in autonomous driving setups. The key ingredient of our method is the use of superpixels to produce object-aware point representation suited for, e.g., semantic segmentation and object detection. We showed that \ours yields powerful point cloud representations which transfer and generalize well to multiple tasks and datasets, surpassing related 
state-of-the-art methods.

\smallparagraph{Acknowledgments.}
We thank Antonin Vobecky, David Hurych, Josef Sivic and Patrick P\'erez for their feedbacks and the fruitful discussions around this project.

%
\cleardoublepage
{\small
\bibliographystyle{ieee_fullname}
\bibliography{egbib}
}

\cleardoublepage
\appendix
\section*{Supplementary Material}

\makeatletter
\newif\ifundottedmtc\undottedmtcfalse
\def\@undottedtocline#1#2#3#4#5{%
\ifnum #1>\c@tocdepth\relax \else
\vskip \z@ plus.2\p@
{\leftskip #2\relax \rightskip \@tocrmarg \parfillskip -\rightskip
\parindent #2\relax\@afterindenttrue
\interlinepenalty\@M
\leavevmode
\@tempdima #3\relax \advance\leftskip \@tempdima \hbox{}%
\hskip -\leftskip
#4\nobreak\hfill \nobreak
\hb@xt@ \@pnumwidth {\hfil \normalfont \normalcolor #5}\par}
\fi}
\makeatother

{
\hypersetup{linkcolor=black}
\startcontents[sections]
\printcontents[sections]{l}{1}{\setcounter{tocdepth}{2}}
}

\makeatletter
\addtocontents{toc}{%
  \protect\begingroup%
  \protect\makeatletter
  \protect\def\protect\l@section{\bf\protect\@undottedtocline{1}{0em}{1.7em}}
  \protect\def\protect\l@subsection{\rm\protect\@dottedtocline{2}{1.5em}{2.4em}}
  \protect\makeatother
}
\makeatother
\vspace{2em}

%
\section{Complementary Results}
\label{sec:more_results}

\subsection{Visual Inspection}

\begin{figure*}[t]
	\centering
	\begin{minipage}{0.48\linewidth}
	    \centering \large Image feature similarity map
	    \includegraphics[width=\linewidth]{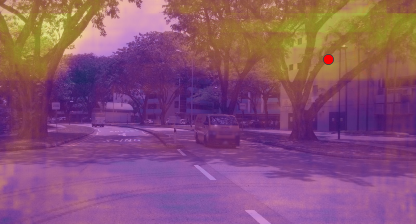}
	\end{minipage}
	\begin{minipage}{0.48\linewidth}
	    \centering \large Point feature similarity map
	    \includegraphics[width=\linewidth]{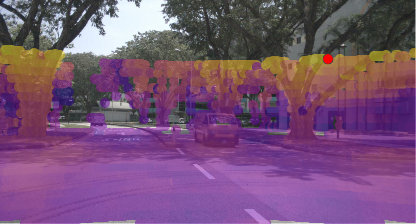}
	\end{minipage}
	\\ \vspace{0.05cm}
	\begin{minipage}{0.48\linewidth}
	    \centering
	    \includegraphics[width=\linewidth]{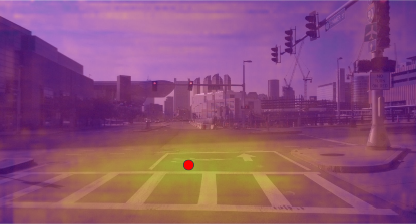}
	\end{minipage}
	\begin{minipage}{0.48\linewidth}
	    \centering
	    \includegraphics[width=\linewidth]{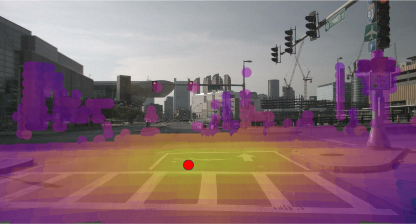}
	\end{minipage}
	\\ \vspace{0.05cm}
	\begin{minipage}{0.48\linewidth}
	    \centering 
	    \includegraphics[width=\linewidth]{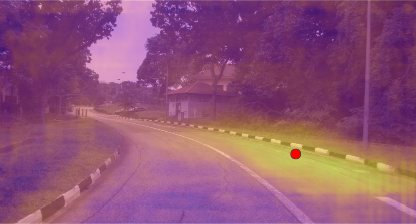}
	\end{minipage}
	\begin{minipage}{0.48\linewidth}
	    \centering 
	    \includegraphics[width=\linewidth]{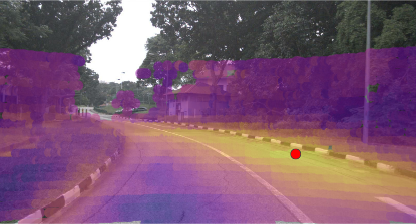}
	\end{minipage}
	\caption{
	Cosine similarity between the \ours's feature of a query point (displayed as a red dot) and: (a) the pixel features of an image in the same scene (left column - image feature similarity map); (b) the features of the other points projected in the same image (right column - point feature similarity map). The colormap goes from violet to yellow for respectively low and high similarity scores. We show these maps for two scenes in the validation set of \cite{nuscenes}.
	}
	\label{fig:feature_maps_supp}
\end{figure*}
\begin{figure*}[t]
	\centering 
	\begin{minipage}{0.48\linewidth}
	    \centering \large Image feature similarity map
	    \includegraphics[width=\linewidth]{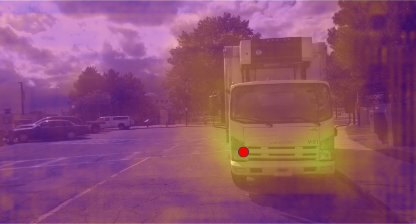}
	\end{minipage}
	\begin{minipage}{0.48\linewidth}
	    \centering \large Point feature similarity map
	    \includegraphics[width=\linewidth]{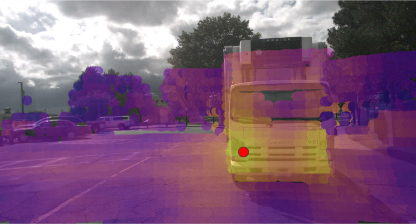}
	\end{minipage}
	\\ \vspace{0.05cm}
	\begin{minipage}{0.48\linewidth}
	    \centering
	    \includegraphics[width=\linewidth]{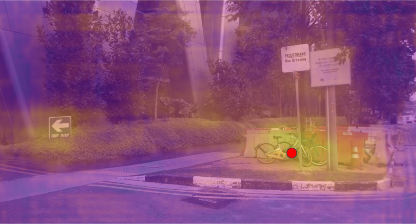}
	\end{minipage}
	\begin{minipage}{0.48\linewidth}
	    \centering
	    \includegraphics[width=\linewidth]{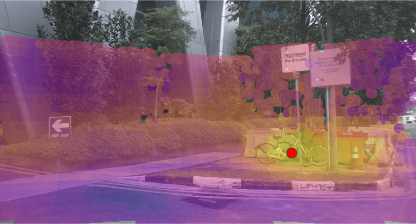}
	\end{minipage}
	\\ \vspace{0.05cm}
	\begin{minipage}{0.48\linewidth}
	    \centering 
	    \includegraphics[width=\linewidth]{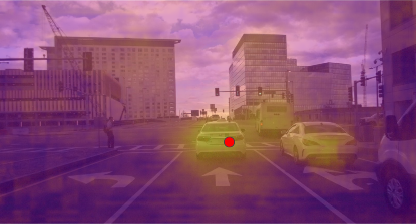}
	\end{minipage}
	\begin{minipage}{0.48\linewidth}
	    \centering 
	    \includegraphics[width=\linewidth]{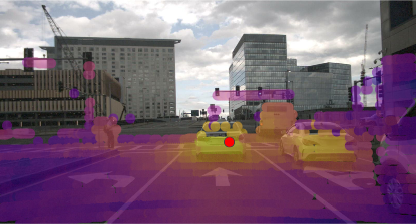}
	\end{minipage}
 	\caption{Cosine similarity between the \ours's feature of a query point (displayed as a red dot) and: (a) the pixel features of an image in the same scene (left column - image feature similarity map); (b) the features of the other points projected in the same image (right column - point feature similarity map). The colormap goes from violet to yellow for respectively low and high similarity scores. We show these maps for two scenes in the validation set of \cite{nuscenes}.}
	\label{fig:feature_maps_supp2}
\end{figure*}

We present in \cref{fig:feature_maps_supp} and \cref{fig:feature_maps_supp2} additional feature similarity maps such as those presented in \cref{fig:feature_maps} in the main paper. We continue to observe the segmentation ability of our pre-trained model: a query point on a tree, road, vehicle, bike is mostly correlated with other points or pixels on trees, road, vehicles, bikes, respectively. 

We also notice some spurious correlations with points or pixels around the objects. These spurious correlations seem more apparent in the image feature similarity maps. This can be due to the reduced resolution of the image feature maps (1/4 in each spatial direction) or that the ResNet-50 features are intrinsically imprecise near object boundaries, which prevent the network to learn accurate near object edges in the 3D point cloud. Increasing the image resolution as well as adding additional constraints leveraging the 3D structures observed in the point cloud can help us to prevent such leakage in future works.

We also provide in the supplementary material a video illustrating the capacity of our pre-trained model in doing semantic segmentation on a sequence of point clouds, without fine-tuning. The video is generated as follows. We choose a scene from the validation set of nuScenes \cite{nuscenes}. We collect all the point features along with the class labels in the first frame of the sequence. This set constitutes our annotated database.
We consider three classes: `vegetation', `car' and `pedestrian'. The points in the following frames are classified by using a binary k-NN classifier (k=20) for each of these class. We display the predicted probability for each class in all subsequent frames: red for `car', blue for `pedestrian', green for `vegetation'.\footnote{A mix of these colors corresponds to predictions with non-zero probability in two or more of these classes.}
We notice that we are able to classify correctly several points in of each of the considered classes throughout the whole sequence. In particular, we detect correctly pedestrians at the beginning of sequence and cars at the end of the sequence. Some points are misclassified but we recall these results are obtained without any fine-tuning of the backbone.

\subsection{Choice of the Image Backbone}

\begin{table}[t]
\centering
\ra{1.1}
\setlength{\tabcolsep}{20pt}
\begin{tabular}{@{}l l c l @{}}
\toprule
    Backbone $g_{\bar{\omega}_{\scriptscriptstyle \text{bck}}}$
        & Pretrain
        & mIoU \\
\midrule
    ResNet-50
        & Full sup.
        & 39.2
    \\
    ResNet-50
        & MoCov2 \cite{chen2020mocov2}
        & 39.2
        \\
    ViT-S/16
        & DINO \cite{caron2021emerging}
        & 39.5
        \\
\bottomrule
\end{tabular}
\vspace{-6pt}
\caption{Performance of \ours on nuScenes semantic segmentation with using different image backbone architectures or pretraining methods. We consider two architectures, ResNet-50 or ViT-S/16, and pretraining under full supervision on ImageNet or by self-supervision (MoCov2, DINO). The scores are obtained by linear probing of the pre-trained backbone. We report the mIoU on our mini-val split.}
\label{tab:image_backbone}
\end{table}

We present in \cref{tab:image_backbone} the performance reached with \ours on nuScenes semantic segmentation when using a ResNet-50 pre-trained under full supervision on ImageNet or under self-supervision using MoCov2. We notice that there is no loss of performance because of the use of self-supervision to pretrain $g_{\bar{\omega}_{\scriptscriptstyle \text{bck}}}$.

We also report in the same table the performance obtained when using a self-supervised transformer \cite{caron2021emerging} as image backbone. The performance is slightly higher than when using a ResNet-50, showing that our method is compatible with this type of image network architectures. \ours can exploit the higher capacity of transformers in learning image representations, which can in turn yields better 3D networks.

\subsection{Choice of the Superpixels Method}

We present in \cref{tab:superpixels_choice} the impact of the choice of the number of superpixels or of the superpixels algorithm on the performance of \ours. 
For Felzenszwalb's \cite{felzenszwalb} method (called FH), we used a scale parameter of $300$, a gaussian pre-processing of standard deviation $0.35$ and a superpixels minimal size of $4000$ pixels, which yield at most 143 superpixels per image on nuScenes' training set. 
For SLIC, we tested 100, 150 or 200 superpixels per image. We see that it is important to adjust the number of superpixels correctly to avoid too much over-segmentation and under-segmentation which both impact negatively the performance. 
The results in the main paper are obtained with SLIC and 150 superpixels per image. 
Finally, we notice that \ours is less sensitive to the choice of superpixel algorithm 
(e.g., using FH instead of SLIC)
once its parameters are set correctly.

\subsection{Few-Shot Semantic Segmentation}

We report in \cref{tab:per_class_semseg_nuscenes} and \cref{tab:per_class_semseg_kitti} the per-class performance of \ours and the different baselines when pretraining on 1$\%$ of the available annotations. We notice that \ppkt and \ours are the two best methods on the majority of the classes with \ours achieving the highest mIoU. On nuScenes, \ours is ranked first on 9 classes vs 6 for \ppkt. On SemanticKITTI, \ours is ranked first on 11 of the classes vs 5 for \ppkt.

%
\section{Data Augmentations}
\label{sec:data_augmentation}

As mentioned in \cref{sec:exp}, we apply two sets of strong data augmentations: the first on point clouds, the second on images. We highlight that implementing these augmentations is not trivial as they impact the list of point-pixel correspondences, which needs to be updated appropriately.

Regarding point clouds, we apply a random rotation around the $z$-axis and flip the direction of the $x$ and $y$-axis with $50\%$ probability for each axis. As in \cite{depthcontrast}, we also drop points that lie in an axis-aligned random cuboid. Concretely, the cuboid center is placed on a randomly-selected point in the point cloud $\ma{P}$ and the length of each side covers at most $10\%$ of the range of point coordinates on the corresponding axis. We make sure that this dropped cuboid preserves at least $1024$ pairs of points and pixels, otherwise another new cuboid is selected.

The images are flipped horizontally $50\%$ of the time, and cropped-resized to $416 \times 224$. The random crop covers at least $30\%$ of the image area with a random aspect ratio between $14/9$ and $17/9$ before resizing. We make sure that this random cropping preserves at least $1024$ or $75\%$ of the pixel-point pairs, otherwise another crop is selected.

%
\section{Baselines' Implementations Details}
\label{sec:baselines_details}

\subsection{PPKT for Autonomous Driving.} 

PPKT \cite{ppkt} was originally proposed for RGB-D data captured indoor. As, up to now, there is no publicly released code, we propose our best adaption of this method in an autonomous driving setup, referred as \ppkt. \ppkt uses the same data augmentations, voxel-based 3D network backbone, and cylindrical coordinate voxels as our method. 

\ppkt is obtained by: using $Q = M$ in \cref{sec:method}, i.e., each superpixel contains one pixel; using strided convolutions in the image backbone 
$g_{\bar{\omega}_{\scriptscriptstyle \text{bck}}}(\cdot)$;
the same image head $h_{\omega_{\scriptscriptstyle \text{head}}}$ as ours but with a bilinear upsamling layer from a resolution $1/32$ in each spatial direction to the original size of the input image. Furthermore, as computing the loss \eqref{eq:distillation_loss} is intractable when considering all possible point-pixel pairs $\set{P} \in \{(\vec{f}^c_{m}, \vec{g}^{c}_{m}) : c=1,\ldots,C, \, m=1,\ldots, M\}$, a random subset of $\set{P}$ is selected to compute it.
In practice, as multiple scenes are available in a batch at train time, the set $\set{P}$ also contains the point-pixel pairs of \emph{all} scenes in our implementation. This sampling is not required in our method thanks to the use of superpixels which reduces the number of matching pairs. We use exactly the same parameters for pre-training a network with \ppkt or \ours.

The noticeable differences between PPKT and \ppkt are the following: 
\begin{enumerate}
    \item the use of cartesian coordinates vs. cylindrical coordinates for $f_\theta$;
    \item direct pixel-point correspondences in RGB-D data vs indirect correspondences computed via a projection matrix in autonomous driving;
    \item the absence of randomly drop cuboids in PPKT and the absence of random rescaling and elastic distorsion in \ppkt.
\end{enumerate}

\begin{table}[t]
\centering
\ra{1.1}
\setlength{\tabcolsep}{7pt}
\begin{tabular}{l l l l l l}
\toprule
    Algorithm
        & \multirow{2}{*}{None}
        & \multicolumn{3}{c}{SLIC \cite{slic}}
        & \multicolumn{1}{c}{FH \cite{felzenszwalb}}
    \\
    \cmidrule(lr){1-1}
    \cmidrule(lr){3-5}
    \cmidrule(lr){6-6}
    \#Superpixels & 
        & 100
        & 150
        & 200
        & $\leq$143
    \\
    \midrule
    mIoU
        & \multicolumn{1}{c}{36.6}
        & 37.7
        & 39.2
        & 36.3
        & \multicolumn{1}{c}{39.2}
    \\
\bottomrule
\end{tabular}
\caption{Sensitivity of \ours to the superpixel algorithms and superpixel parameters. The semantic segmentation scores (mIoU) are obtained by linear probing and computed on our nuScenes mini-val split. We compare the performance of \ours when using (a) no superpixels; (b) SLIC \cite{slic} with different number of superpixels per image; and (c) the Felzenszwalb's \cite{felzenszwalb} algorithm (FH) with parameters that produce at most 143 superpixels per image.}
\label{tab:superpixels_choice}
\end{table}
%

\begin{table*}[t]
\centering
\newcommand*\rotext{\multicolumn{1}{R{60}{1em}}}
\setlength{\tabcolsep}{4pt}
\begin{tabular}{@{}l l l l l l l l l l l l l l l l l | l}
\toprule
    Method
    & \rotext{barrier}
    & \rotext{bicycle}
    & \rotext{bus}
    & \rotext{car}
    & \rotext{const. veh.}
    & \rotext{motorcycle}
    & \rotext{pedestrian}
    & \rotext{traffic cone}
    & \rotext{trailer}
    & \rotext{truck}
    & \rotext{driv. surf.}
    & \rotext{other flat}
    & \rotext{sidewalk}
    & \rotext{terrain}
    & \rotext{manmade}
    & \rotext{vegetation}
    & \rotext{\bf mIoU}
    \\
\midrule
    Random 
        & 0.0
        & 0.0
        & \textcolor{white}{0}8.1
        & 65.0
        & 0.1
        & \textcolor{white}{0}6.6
        & 21.0
        & \textcolor{white}{0}9.0
        & \textcolor{white}{0}9.3
        & 25.8
        & 89.5
        & 14.8
        & 41.7
        & 48.7
        & 72.4
        & 73.3
        & 30.3
    \\
    \pointc
        & 0.0
        & 1.0
        & \textcolor{white}{0}5.6
        & 67.4
        & 0.0
        & \textcolor{white}{0}3.3
        & 31.6
        & \textcolor{white}{0}5.6
        & 12.1
        & 30.8
        & 91.7
        & 21.9
        & 48.4
        & 50.8
        & 75.0
        & 74.6
        & 32.5
    \\  
    \depthc  
        & 0.0
        & 0.6
        & \textcolor{white}{0}6.5
        & 64.7
        & 0.2
        & \textcolor{white}{0}5.1
        & 29.0
        & \cellcolor{second}\textcolor{second}{0}9.5
        & 12.1
        & 29.9
        & 90.3
        & 17.8
        & 44.4
        & 49.5
        & 73.5
        & 74.0
        & 31.7
    \\
    \ppkt 
        & 0.0
        & \cellcolor{second}2.2
        & \cellcolor{first}20.7
        & \cellcolor{first}75.4
        & \cellcolor{first}1.2
        & \cellcolor{second}13.2
        & \cellcolor{first}45.6
        & \textcolor{white}{0}8.5
        & \cellcolor{first}17.5
        & \cellcolor{first}38.4
        & \cellcolor{second}92.5
        & \cellcolor{second}19.2
        & \cellcolor{second}52.3
        & \cellcolor{second}56.8
        & \cellcolor{second}80.1
        & \cellcolor{second}80.9
        & \cellcolor{second}37.8
    \\
    \ours 
        & 0.0
        & \cellcolor{first}3.1
        & \cellcolor{second}15.2
        & \cellcolor{second}72.0
        & \cellcolor{second}0.9
        & \cellcolor{first}18.8
        & \cellcolor{second}43.2
        & \cellcolor{first}12.5
        & \cellcolor{second}14.7
        & \cellcolor{second}33.3
        & \cellcolor{first}92.8
        & \cellcolor{first}29.4
        & \cellcolor{first}54.0
        & \cellcolor{first}61.0
        & \cellcolor{first}80.2
        & \cellcolor{first}81.9
        & \cellcolor{first}38.3
    \\
\bottomrule
\end{tabular}
\caption{Per-class performance on nuScenes using $1\%$ of the annotated scans for fine-tuning. We report the IoU for each class and highlight the best and second best scores with dark blue and light blue backgrounds, respectively.}
\label{tab:per_class_semseg_nuscenes}
\end{table*}

\begin{table*}[t]
\centering
\newcommand*\rotext{\multicolumn{1}{R{60}{1em}}}
\setlength{\tabcolsep}{2.3pt}
\begin{tabular}{@{}l l l l l l l l l l l l l l l l l l l l | l}
\toprule
    Method
    & \rotext{car}
    & \rotext{bicycle}
    & \rotext{motorcycle}
    & \rotext{truck}
    & \rotext{other-vehicle}
    & \rotext{person}
    & \rotext{bicyclist}
    & \rotext{motorcyclist}
    & \rotext{road}
    & \rotext{parking}
    & \rotext{sidewalk}
    & \rotext{other-ground}
    & \rotext{building}
    & \rotext{fence}
    & \rotext{vegetation}
    & \rotext{trunk}
    & \rotext{terrain}
    & \rotext{pole}
    & \rotext{traffic-sign}
    & \rotext{\bf mIoU}
    \\
\midrule
    Random   
        & 91.2
        & \textcolor{white}{0}0.0
        & \textcolor{white}{0}9.4
        & \textcolor{white}{0}8.0
        & 10.7
        & 21.2
        & \textcolor{white}{0}0.0
        & \textcolor{white}{0}0.0
        & 89.4
        & 21.4
        & 73.0
        & \cellcolor{first}\textcolor{first}{0}1.1
        & 85.3
        & \cellcolor{second}41.1
        & 84.9
        & 50.1
        & 71.4
        & 55.4
        & 37.6
        & 39.5
    \\
    \pointc  
        & 90.1
        & \cellcolor{second}\textcolor{second}{0}4.6
        & \textcolor{white}{0}5.4
        & \textcolor{white}{0}8.1
        & \textcolor{white}{0}9.5
        & 21.9
        & \cellcolor{second}30.8
        & \textcolor{white}{0}0.0
        & 90.7
        & 25.6
        & 73.3
        & \textcolor{white}{0}0.3
        & 86.4
        & 39.3
        & 83.7
        & 51.2
        & 70.6
        & 53.6
        & 34.9
        & 41.1
    \\  
    \depthc  
        & \cellcolor{second}91.7
        & \cellcolor{first}\textcolor{first}{0}8.8
        & \cellcolor{second}11.5
        & 19.9
        & \cellcolor{first}15.4
        & 24.7
        & \textcolor{white}{0}0.0
        & \textcolor{white}{0}0.0
        & 89.5
        & 21.0
        & 72.9
        & \cellcolor{second}\textcolor{second}{0}0.7
        & 85.4
        & 40.6
        & \cellcolor{second}85.1
        & 51.7
        & \cellcolor{second}71.3
        & \cellcolor{second}57.9
        & 39.3
        & 41.5
    \\
    \ppkt 
        & 91.3
        & \textcolor{white}{0}1.9
        & 11.2
        & \cellcolor{first}23.1
        & 12.1
        & \cellcolor{second}27.4
        & \cellcolor{first}37.3
        & \textcolor{white}{0}0.0
        & \cellcolor{first}91.3
        & \cellcolor{second}27.0
        & \cellcolor{second}74.6
        & \textcolor{white}{0}0.3
        & \cellcolor{second}86.5
        & 38.2
        & \cellcolor{first}85.3
        & \cellcolor{second}58.2
        & \cellcolor{first}71.6
        & 57.7
        & \cellcolor{second}40.1 
        & \cellcolor{second}43.9
    \\
    \ours  
        & \cellcolor{first}92.2
        & \textcolor{white}{0}3.0
        & \cellcolor{first}17.0
        & \cellcolor{second}22.4
        & \cellcolor{second}14.3
        & \cellcolor{first}36.0
        & 22.1
        & \textcolor{white}{0}0.0
        & \cellcolor{first}91.3
        & \cellcolor{first}30.0
        & \cellcolor{first}74.7
        & \textcolor{white}{0}0.2
        & \cellcolor{first}87.7
        & \cellcolor{first}41.2
        & 85.0
        & \cellcolor{first}58.5
        & 70.4
        & \cellcolor{first}58.3
        & \cellcolor{first}42.4
        & \cellcolor{first}44.6
    \\
\bottomrule
\end{tabular}
\caption{Per-class performance on SemanticKITTI using $1\%$ of the annotated scans for fine-tuning. We report the IoU for each class and highlight the best and second best scores with dark blue and light blue backgrounds, respectively.}
\label{tab:per_class_semseg_kitti}
\end{table*}

\subsection{PointContrast} 

We retrained PointContrast \cite{pointcontrast} on nuScenes after studying several setups to optimize its performance. PointContrast requires pairs of point clouds acquired from different viewpoints in the same scene with a list of matching points in these two views. To provide a fair baseline, we tested different strategies to create this training dataset. Among the tested strategies, the best one consists in creating all possible pairs of keyframes within a scene, then removing pairs of point clouds which are less than $10$ m apart, and removing those which have less than $1024$ pairs of matching points. 

The list of matching points in a pair of point clouds is computed as follows. We first register both point clouds using the ground truth pose of the Lidar. Then, for each point in one point cloud, we search for the nearest point in the second point cloud and consider that it is a pair of matching points if the points are less than $10$ cm apart. 

\pointc is trained on $1$ GPU, with a batch size of $8$, using SGD with the same parameters as for \ours, except for a initial rate set at $1$, and cosine annealing scheduler. We selected the learning rate using our mini-val split in order to optimize the performance of \pointc. As point cloud augmentations, we used a random rotation around the z-axis, random flip of the x or y-axis and dropped points in cuboids whose sides cover at most $20\%$ of the range of point coordinates in each axis. Finally, one can note that the size of the pre-training dataset is different for \pointc and \ours. For fairness, we set the number of iterations for pre-training with \pointc as follows: we compute the total number of point clouds used in \ours over the $50$ training epochs and use the same number of pairs of point clouds in \pointc.

\subsection{DepthContrast} 

The last baseline is DepthContrast \cite{depthcontrast} which pre-trains simultaneously two 3D network backbones, 
a point-based network, e.g., \cite{pointnet2} and a voxel-based network, e.g., \cite{voxelnet}, using a contrastive task between the global point-cloud representations of the two networks.
Among the three baselines, it is the only one which has already been used on a autonomous driving dataset: the Waymo Open Dataset \cite{waymo}. 
To make it comparable with the rest of the methods in our study, we re-used the point-based network used in \cite{depthcontrast} while changing the voxel-based network to the same sparse residual U-Net that processes cylindrical coordinate voxels as in \ours. 
After DepthContrast pre-training, we only evaluate on the downstream tasks the voxel-based network.

We trained \depthc on $1$ GPU, with a batch size of $8$, using SGD with a momentum of $0.9$, weight decay of $0.0001$, and an initial learning rate of $0.001$ (tuned on our mini-val split) that drops to $10^{-6}$ with a cosine annealing scheduler. We used queues of $60{\rm K}$ negatives for the contrastive loss.

%
\section{Additional Training Details}
\label{sec:training_details}

\subsection{Linear Probing}

In this experiment, the pretrained network $f_\theta$ is combined with a pointwise linear classification head which is trained for $50$ epochs with a learning rate of $0.05$ for all methods.

\subsection{Few-Shot Semantic Segmentation}

On SemanticKITTI, the networks are fine-tuned for $100$ epochs and a batch size of $10$. On nuScenes, we use a batch size of $16$ and for $100$ epochs. We use different learning rates on the classification head and the backbone $f_\theta$, except when the backbone is initialized with random weights. We recall that these learning rates are optimized for each method and each dataset, using our mini-val split for nuScenes and the validation set for semanticKITTI. 

\subsection{Annotation Efficiency on nuScenes}

As in the previous section, we use different learning rates on the classification head and the backbone $f_\theta$ and optimize these learning rates for each method and each subset size, using our mini-val split. The network is fine-tuned for $100$ epochs when using $1\%$ of annotated data and $50$ epochs for the other percentages.

%
\section{nuScenes mini-val Split}
\label{sec:mini_val}

The training set of nuScenes contains $700$ scenes in total, including:
\begin{itemize}[itemsep=-1pt,topsep=-1pt]
    \item 137 raining scenes,
    \item 84 night-time scenes,
    \item 310 scenes in Singapore,
    \item 390 scenes in Boston.
    \vspace{1mm}
\end{itemize}

We construct our mini-val split by selecting $100$ scenes from this training set so that it contains each type of scenes in the same proportion. Our mini-val split contains:
\begin{itemize}[itemsep=-1pt,topsep=-1pt]
    \item 20 raining scenes,
    \item 12 night-time scenes,
    \item 44 scenes in Singapore,
    \item 56 scenes in Boston.
    \vspace{1mm}
\end{itemize}
We will provide the list of selected scenes along with the implementation of \ours.

\section{Societal and Environmental Impact}

Self-supervision enables the use of large and uncurated datasets with performance often increasing with the duration of the training schedule and the size of the model, at the cost of using much more computational resources with possibly negative environmental impacts. Yet, pre-trained models also reduce the training time needed on multiple downstream tasks, hence reducing the environmental cost of training downstream models.
In fact, we distribute our pre-trained models.

\section{Public Resources Used}

We acknowledge the use of the following public resources, during the course of this work:

\begin{itemize}[itemsep=-1pt,topsep=-1pt]
    \item KITTI object detection    \dotfill CC BY-NC-SA 3.0
    \item MinkowskiEngine           \dotfill MIT License
    \item nuScenes                  \dotfill CC BY-NC-SA 4.0
    \item nuScenes-devkit           \dotfill Apache License 2.0
    \item OpenPCDet                 \dotfill Apache License 2.0
    \item Pytorch Lightning         \dotfill Apache License 2.0
    \item Semantic KITTI            \dotfill CC BY-NC-SA 4.0
\end{itemize}

\addtocontents{toc}{%
  \protect\endgroup%
}


\end{document}